\documentclass[sigconf]{acmart}

\usepackage{multirow}
\usepackage{boldline}
\usepackage{makecell}
\copyrightyear{2022} 
\acmYear{2022} 
\setcopyright{acmlicensed}\acmConference[WebSci '22]{14th ACM Web Science Conference 2022}{June 26--29, 2022}{Barcelona, Spain}
\acmBooktitle{14th ACM Web Science Conference 2022 (WebSci '22), June 26--29, 2022, Barcelona, Spain}
\acmPrice{15.00}
\acmDOI{10.1145/3501247.3531566}
\acmISBN{978-1-4503-9191-7/22/06}

\acmConference[WebSci '22]{14th International ACM Conference on Web Science in 2022 }{June 26--29,
  2022}{Barcelona, Spain}
\begin{document}


\title[Problem of Semantic Shift]{The Problem of Semantic Shift in\\ Longitudinal Monitoring of Social Media}
\subtitle{A Case Study on Mental Health During the COVID-19 Pandemic}


\author{Keith Harrigian}
\email{kharrigian@jhu.edu}
\orcid{0000-0001-9483-8469}
\affiliation{%
  \institution{Johns Hopkins University}
  \city{Baltimore}
  \state{Maryland}
  \country{USA}
}

\author{Mark Dredze}
\email{mdredze@cs.jhu.edu}
\orcid{0000-0002-0422-2474}
\affiliation{%
  \institution{Johns Hopkins University}
  \city{Baltimore}
  \state{Maryland}
  \country{USA}
}

\renewcommand{\shortauthors}{Harrigian and Dredze}


\begin{abstract}
Social media allows researchers to track societal and cultural changes over time based on language analysis tools. Many of these tools rely on statistical algorithms which need to be tuned to specific types of language. Recent studies have shown the absence of appropriate tuning, specifically in the presence of semantic shift, can hinder robustness of the underlying methods. However, little is known about the practical effect this sensitivity may have on downstream longitudinal analyses. We explore this gap in the literature through a timely case study: understanding shifts in depression during the course of the COVID-19 pandemic. We find that inclusion of only a small number of semantically-unstable features can promote significant changes in longitudinal estimates of our target outcome. At the same time, we demonstrate that a recently-introduced method for measuring semantic shift may be used to proactively identify failure points of language-based models and, in turn, improve predictive generalization.
\end{abstract}

\ccsdesc[300]{Information systems~Social networks}
\ccsdesc[300]{Applied computing~Health care information systems}
\ccsdesc[300]{Computing methodologies~Semi-supervised learning settings}
\ccsdesc[300]{Computing methodologies~Modeling methodologies}


\keywords{semantic shift, longitudinal monitoring, mental health}


\maketitle


\section{Introduction} \label{sec:Introduction}

Across multiple disciplines, studies of social media text and metadata have yielded valuable insights into population-level dynamics (e.g., consumer habits \cite{saura2019black}, voting patterns \cite{beauchamp2017predicting}). In several cases, the outcomes have enabled policy makers to more effectively anticipate and respond to concerns amongst their constituents \cite{myslin2013using,burnap2015cyber}. Now, as the world is presented with new and evolving global crises -- e.g., COVID-19, climate change, and racial inequity -- researchers look to build upon the utility of these past analyses to inform decision-making that is almost certain to have enduring ramifications \cite{van2021artificial}. 

Methods based on machine learning (ML), natural language processing (NLP), and web mining build the foundation of these efforts, offering an opportunity to answer questions that cannot be easily addressed using traditional mechanisms alone \cite{paul2016social,nobles2018std}. At the same time, researchers in these computational communities are aware of how brittle these methods can be. The challenges of transfer learning and domain adaptation are well known \cite{blitzer2007biographies,ruder2019transfer}, with various algorithmic techniques having since been developed to enhance model robustness and improve generalization within novel data distributions \cite{jiang2007instance,huang2019neural}. Yet, how these problems and their proposed solutions affect conclusions within longitudinal studies remains largely absent from applied analyses.

Indeed, longitudinal studies almost ubiquitously follow the same formulaic approach. First, acquire ground truth for a target concept within a small sample of data (e.g., regular expressions to identify medical diagnosis disclosures \cite{coppersmith2014quantifying}, follower networks indicating political leaning \cite{al2012homophily}). Next, train a statistical classifier on this data with the objective of re-identifying language associated with the target concept. Finally, apply the trained classifier to a new population of individuals across multiple time steps (e.g., annually, weekly). The first two stages of this modeling procedure have been explored extensively \cite{volkova2015inferring}, but studies validating the final step have been sparse, due largely to the difficulties of obtaining temporally granular ground truth for many high-level concepts \cite{zafarani2015evaluation,choi2020development}.

A lack of analyses of temporal robustness of these models belies the seriousness of the problem: language shifts over time -- especially on social media \cite{brigadir2015analyzing,loureiro2022timelms} -- and statistical classifiers degrade in the presence of distributional changes \cite{daume2006domain,huang2019neural}. Three types of distributional change are of particular concern for classifiers applied over time: 1) new terminology is used to convey existing concepts; 2) existing terminology is used to convey new concepts; and 3) semantic relationships remain fixed, but the overall language distribution changes. The latter challenge frequently manifests when major social events cause large-scale shifts in the topic of online conversation (e.g., discussion of healthcare increases during a pandemic, discussion of a political leader increases near an election). Unfortunately, these are often the types of events we seek to study.

To better understand this gap in the literature, we conduct a case study on estimating changes in depression prevalence during the COVID-19 pandemic, a timely analysis with value to the medical and public health communities which has thus far has procured incongruous results across studies \cite{galea2020mental,bray2021racial}. We draw inspiration from research on detecting distributional shifts in language over time \cite{dredze2010we,huang2018examining}, focusing our attention on a recently-introduced method that leverages word embedding neighborhoods to identify semantic shift between multiple domains \cite{gonen2020simple}. We find that semantically-informed feature selection can improve classifier generalization when semantic noise and predictive power are interwoven. More importantly, we provide evidence that semantic shift can introduce undesirable variance in downstream longitudinal monitoring applications, despite having an indistinguishable effect on historical predictive performance. Altogether, our study serves as a cautionary tale to practitioners interested in using social media data and statistical algorithms to derive sensitive population insights.


\section{Motivation} \label{sec:Motivation}

When the COVID-19 pandemic began in March 2020, healthcare professionals warned of an impending mental health crisis, with economic uncertainty \cite{godinic2020effects}, loss of access to care \cite{yao2020patients}, and physical distancing \cite{galea2020mental} expected to reduce mental wellness. Given the inherent difficulties of measuring mental health at scale using traditional monitoring mechanisms, the healthcare community called upon computational scientists to leverage web data to provide evidence for optimizing crisis mitigation strategies \cite{torous2020digital}. Computational researchers responded by analyzing search queries regarding anxiety and suicidal ideation \cite{ayers2020internet,ayers2021suicide}, developing novel topic models to gather an understanding of the population's concerns \cite{koh2020loneliness}, and applying language-based classifiers to streams of social media text \cite{wolohan2020estimating}.

Unfortunately, these inquiries failed to provide unanimous insights that could be used with any confidence to manage the ongoing situation \cite{zelner2021accounting}. For instance, application of a neural language classifier to the general Reddit population estimated over a 50\% increase in depression after the start of the pandemic \cite{wolohan2020estimating}, despite an analysis of topic distributions within three mental health support subreddits finding evidence to suggest the opposite \cite{biester2020quantifying}. Similarly, multiple keyword-based analyses using Google Trends data suggested anxiety increased relative to expected levels \cite{stijelja2020covid,ayers2020internet}, while others suggested anxiety levels actually remained stable \cite{knipe2020mapping}.

Two years later, our understanding of COVID-19's effect on mental health is still evolving. For example, in a survey conducted early in the pandemic by the Centers for Disease Control and Prevention (CDC), 10.7\% of respondents reported having thoughts of suicide in the previous 30 days \citep{czeisler2020mental} (a $2\times$ increase over the expected rate). Later in 2020, data suggested suicide rates remained stable or even fell after the start of the COVID-19 pandemic \citep{ahmad2021quarterly}. Some argued this drop was the result of a ``pulling together'' effect \citep{ayers2021suicide}, an outcome that had been observed previously during times of crisis \cite{claassen2010effect,gordon2011impact}. However, upon closer inspection, it became clear that this trend was subject to Simpson's paradox \citep{julious1994confounding}. Reductions in suicide rate were observed amongst white folk, while a significant increase was observed amongst ethnic and racial minorities \citep{mcknight2021racial}, with whom stress-inducing factors such as financial instability and food insecurity are more common. These nuances illustrate one dimension of difficulty associated with monitoring mental health -- heterogeneity. In this paper, we will highlight an understudied dimension that affects analyses of web data -- semantic shift.

\begin{figure}
    \centering
    \includegraphics[width=\columnwidth]{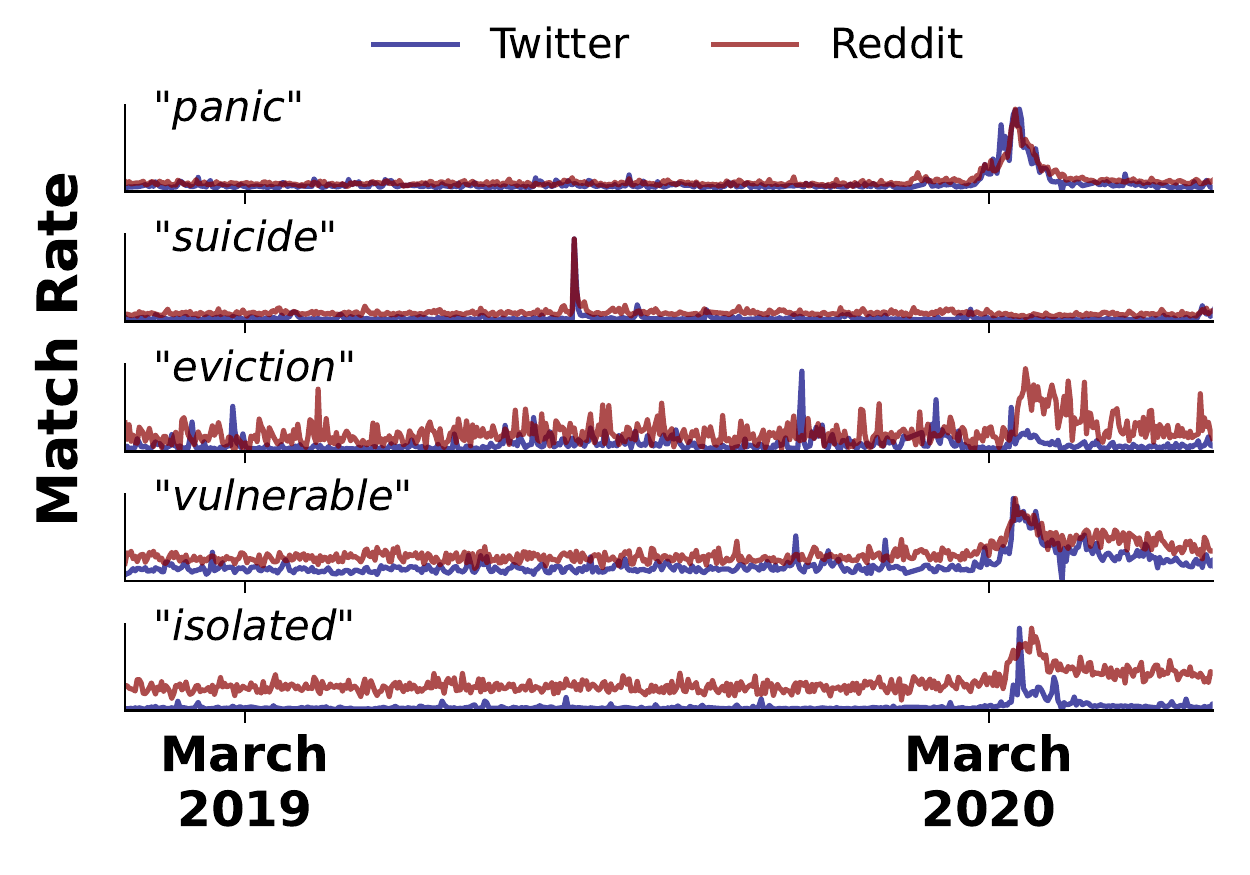}
    \caption{Proportion of posts containing a subset of depression-indicative $n$-grams.}
    \label{fig:KeywordTimeseries}
\end{figure}

\subsection{Understanding the Uncertainty} \label{sec:understand}

In general, it is common for web-based research utilizing different datasets and analysis techniques to arrive at varying measurements of a specific phenomenon \cite{hutton2015toward,roy2018clean}. However, failure to understand \emph{why} these discrepancies emerge critically constrains our ability to instill confidence in future use of computational monitoring methods. In the case of the aforementioned computational studies, we argue the primary distinction is the manner in which linguistic units are aggregated and transformed into downstream insights.

Indeed, a review of mental-health related $n$-gram usage\footnote{Pointwise Mutual Information of each term within historical samples of depressed individuals was used to determine mental health relevance.} over time (and inspection of the posts in which they are found) highlights how inclusion of semantically unstable terms could confound results (see Figure \ref{fig:KeywordTimeseries}). For example, spikes in usage of the term ``suicide'' in August 2019 are actually a response to Jeffrey Epstein's death. Meanwhile, increased usage of ``panic,'' ``eviction,'' ``vulnerable,'' and ``isolated'' in March 2020 primarily corresponds to discussion of pandemic-specific circumstances (e.g., toilet paper \emph{panic}, \emph{eviction} moratorium, medically \emph{vulnerable} populations).

Keyword-based monitoring methods such as those from \citet{stijelja2020covid}, which aggregate counts from a pool of several lexical units together, are vulnerable to measurement error if the underlying semantic distribution of any subset of terms changed during the course of the given analysis period. Statistical language models have a higher capacity to disambiguate contextual usage by leveraging thousands of lexical units simultaneously to arrive at a final inference \cite{biester2020quantifying,wolohan2020estimating}. However, they provide no guarantee that this disambiguation is done correctly in the presence of dramatic distributional shifts \cite{dhingra2021time}. These challenges raise two important questions: 

\begin{enumerate}
    \item To what extent has semantic shift affected results regarding mental health in the published literature?
    \item Is it possible to obtain more reliable longitudinal estimates by explicitly invoking knowledge of semantic shift when training statistical algorithms?
\end{enumerate}



\section{Related Work} \label{sec:related}

In the machine learning literature, semantic shift often induces what is known as \emph{concept shift} \citep{widmer1996learning,lu2018learning}. That is, we experience a mismatch between the training and deployment conditional distributions $p(y \mid x)$. These shifts occur naturally in a variety of situations where time is involved \citep{ruano2018concept,hu2020no}. For example, consider the word ``lonely'' becoming a weaker indicator of depression  during Valentine's day when loneliness is expressed more frequently in a romantic context.

Concept shift is often addressed in a two-stage ``detect, then react'' manner \citep{yang2005combining}. During the detection phase, the goal is to measure similarity between source and target distributions, raising a flag when the divergence surpasses a predetermined threshold \citep{vallim2014proposal,webb2018analyzing}. When representative labels are available for the target distribution, we can simply select a mixture of source and target data to train a new, more appropriate model \citep{vorburger2006entropy}. In the much more common case when labels for the target distribution are not available, methods attempt to identify subsets of source data that contain patterns found within the target data \citep{hulten2001mining} or build an ensemble of experts weighted by their appropriateness to the target distribution \citep{last2002online,kolter2007dynamic}. Alternative methods operate on the target predictions instead, applying post-hoc corrections to re-calibrate the estimated probabilities \citep{tian2021exploring}. Our study is more similar to the former; we use unlabeled deployment data to inform feature selection. 


\section{Measuring Semantic Shift} \label{sec:Measuring}

The type of manual lexical analysis discussed in \S \ref{sec:understand} is not feasible to perform at scale for statistical language models that often have vocabularies with thousands of terms. Fortunately, a substantial pool of prior work has proposed methods for algorithmically quantifying semantic shift between language domains \cite{dredze2010we,kutuzov2018diachronic}. We choose to leverage a method introduced recently by \citet{gonen2020simple}, which has not only outperformed several state-of-the-art alternatives in preliminary studies \cite{hamilton2016diachronic}, but also shown promise for use by applied practitioners. Core advantages of this methodology include its interpretability, robustness to stochasticity, ease of implementation, and low computational overhead.

\citeauthor{gonen2020simple}'s method \cite{gonen2020simple} assumes that semantically stable language has similar sets of neighboring lexical units within word embedding spaces of different domains. More formally, for two domains $\mathcal{P}$ and $\mathcal{Q}$, the semantic stability $S$ of a lexical unit (e.g., word, $n$-gram) $w$ can be measured as:
\begin{align*}
    S(w; \mathcal{P}, \mathcal{Q}) = \frac{\text{nb}_{\mathcal{P}}^{(k)}(w) \cap \text{nb}_{\mathcal{Q}}^{(k)}(w)}{k},
\end{align*}
where $\text{nb}_{\mathcal{X}}^{(k)}(w)$ (i.e., the neighborhood of $w$ in $\mathcal{X}$) denotes the top-$k$ set of lexical units nearest to lexical unit $w$ in word-embedding vector space $X$ based on a vector distance metric of the modeler's choosing. Hyperparameters include the neighborhood size $k$, the minimum frequency of $n$-grams used for building each neighborhood $\text{cf}_{\text{nb}}$, the minimum frequency of $n$-grams input to the semantic shift calculation $\text{cf}_{\text{shift}}$, the distance function used for measuring a word's neighborhood, and the embedding model architecture. For the purpose of measuring semantic shift longitudinally, we can think of independent, discrete time periods as the domains $\mathcal{P, Q}$. We include our parameterization choices in Appendix \ref{apx:predparam}.


\begin{table}[t!]
\centering
\begin{tabular}{ccccc}
\textbf{Dataset} & 
    \textbf{Platform} & 
    \textbf{Dates} & 
    \textbf{\# Users} \\ \toprule
CLPsych \cite{coppersmith2015clpsych} &
    Twitter &
    2012 - 2014 &
    \begin{tabular}{@{}c@{}}C: 477\\D: 477\end{tabular} \\ 
\begin{tabular}{@{}c@{}}Multi-Task\\Learning \cite{benton2017multitask}\end{tabular} &
    Twitter &
    2012 - 2016 &
    \begin{tabular}{@{}c@{}}C: 1,400\\D: 1,400\end{tabular} \\ 
SMHD \cite{cohan2018smhd}  &
    Reddit &
    2013 - 2018 &
    \begin{tabular}{@{}c@{}}C: 127,251\\D: 14,139\end{tabular}\\ 
\begin{tabular}{@{}c@{}}Topic-Restricted\\Text \cite{wolohan2018detecting}\end{tabular} &
    Reddit &
    2016 - 2020 &
    \begin{tabular}{@{}c@{}}C: 107,274\\D: 9,210\end{tabular} \\ \midrule
1\% Stream &
    Twitter &
    1/2019 - 7/2020 &
    All: 25,379 \\
Pushshift.io \cite{baumgartner2020pushshift} &
    Reddit &
    1/2019 - 7/2020 &
    All: 40,671 \\ \bottomrule
\end{tabular}
\caption{Summary statistics for labeled (top) and unlabeled (bottom) datasets. Labeled dataset statistics are further broken out as a function of (C)ontrol and (D)epression groups.}  \label{tab:DataTable}
\end{table}

\section{Data} \label{sec:Data}

To comprehensively understand how semantic shift may influence downstream longitudinal analyses, we leverage datasets which come from multiple social media platforms, span a wide range of time periods, and leverage different annotation/sampling mechanisms. As mentioned before (\S \ref{sec:Motivation}), we focus on the the task of estimating depression prevalence, an important undertaking within the public health community \cite{gelenberg2010prevalence} due to the substantial burden on individuals, communities, and society \cite{dressler1991stress, lepine2011increasing,chang2012economic}. While we consider this specific use case, our analyses are generally applicable to longitudinal monitoring of social media.

\textbf{Institutional Oversight} This research was deemed exempt from review by our Institutional Review Board (IRB) under 45 CFR \S 46.104. All datasets in this study are either publicly accessible (via authenticated application programming interfaces) or available through secure distribution mechanisms (i.e., non-commercial data usage agreements). Given the sensitive nature of mental health data, we abide by additional protocols enumerated in \citet{benton2017ethical} to govern all data collection, storage, and analysis. We discuss the ethics of conducting this type of research (i.e., sensitive attribute tracking) in \S \ref{sec:Ethics}.

\subsection{Labeled Data}

To numerically quantify the effect semantic shift has on predictive generalization, we consider four widely adopted datasets containing ground truth annotations of individual-level depression status. To diversify our data sample and understand platform-specific differences, we consider two Twitter datasets -- 2015 CLPsych Shared Task \cite{coppersmith2015clpsych}, Multi-Task Learning \cite{benton2017multitask} -- and two Reddit datasets -- Topic-restricted Text \cite{wolohan2018detecting}, Self-reported Mental Health Diagnoses (SMHD) \cite{cohan2018smhd}.  Each dataset relies on a form of distant supervision; the Topic-restricted Text dataset assumes original posts made in the r/depression subreddit serve as a proxy for a depression diagnosis, while the remaining three datasets use regular expressions to identify self-disclosures of a depression diagnosis. This annotation procedure remains widely used to train classifiers for monitoring population-level trends due to challenges inherent in acquiring sufficient samples of annotated data \cite{de2013social,chancellor2020methods}, but remains prone to introducing label noise and other sampling artifacts \cite{ernala2019methodological}.


\subsection{Unlabeled Data}


Our primary interest is understanding the practical effects of semantic shift in longitudinal monitoring applications. Accordingly, we collect large samples of text data from both Twitter and Reddit to use for extrinsic model evaluation. Our sampling procedures are inspired by those used in prior COVID-19 related work \citep{saha2020psychosocial,wolohan2020estimating}.

We acquire raw Twitter data from the platform's streaming API, a 1\% sample of all public tweets available for non-commercial research use. We isolate all original tweets (i.e., no retweets) that include an `en' language metadata attribute and are further classified as being written in English based on automatic language identification \cite{lui2012langid}. To facilitate application of our statistical classifiers, which require multiple documents from each individual to make accurate inferences, we further isolate individuals with at least 400 posts across the entire study time period (January 1, 2019 through July 1, 2020).

We sample Reddit data within the same time period using the Pushshift.io archive \cite{baumgartner2020pushshift}, which, unlike the Twitter streaming API, provides access to nearly all historical Reddit data \cite{gaffney2018caveat}. We begin data collection by identifying all users who posted a comment in one of the 50 most popular subreddits\footnote{Based on total number of subscribers as of 6/01/2020. Statistics sourced from https://subredditstats.com} between May 25, 2020 and June 1, 2020. Of the 1.2 million unique users identified by this query, roughly 200k were identified to have posted at least once per week during January 2019 and to not exhibit clear indicators of bot activity (e.g., repeated comments, username indicators, abnormal activity volume). We collect the entire public comment history from January 1, 2019 through July 1, 2020 for a random sample of 50k users in this cohort and perform additional filtering to isolate English data and users who have at least 200 posts across the study time period. Summary statistics for all datasets are provided in Table \ref{tab:DataTable}.



\begin{table*}[t]
    \centering
    \begin{tabular}{c c c c c c c c c c c}
    & & & & \multicolumn{3}{c}{\textbf{Na\"{i}ve}} & \multicolumn{2}{c}{\textbf{Statistical}} & \multicolumn{2}{c}{\textbf{Semantic}} \\ \cmidrule(lr){5-7} \cmidrule(lr){8-9} \cmidrule(lr){10-11}
          &
          \textbf{Dataset} & 
          \textbf{Train} &
          \textbf{Test} &
          \textbf{Cumulative} &
          \textbf{Intersection} &
          \textbf{Frequency} &
          \textbf{Chi-Squared} &
          \textbf{Coefficient} &
          \textbf{Overlap} &
          \textbf{Weighted} \\ \toprule
    \multirow{7}{*}{\rotatebox[origin=c]{90}{\textbf{Twitter}}} &
        CLPysch & 
            2012-2013 & 
                2013-2014 & 
                    0.656\hphantom{*} &
                    0.677\hphantom{*} &
                    0.676\hphantom{*} &
                    0.687\hphantom{*} &
                    0.677\hphantom{*} &
                    \textbf{0.715*} &
                    0.696\hphantom{*} \\ \cmidrule{2-11}
        & \multirow{2}{*}{\begin{tabular}{@{}c@{}}Multi-Task\\Learning\end{tabular}} & 
            2012-2013 &
                2013-2014  & 
                    0.746\hphantom{*} &
                    0.759\hphantom{*} &
                    0.759\hphantom{*} &
                    0.761\hphantom{*} &
                    0.757\hphantom{*} &
                    \textbf{0.779*} &
                    0.772\hphantom{*}\\
            & & &
                2014-2015  & 
                    0.703\hphantom{*} &
                    0.760\hphantom{*} &
                    0.760\hphantom{*} &
                    0.762\hphantom{*} &
                    0.758\hphantom{*} &
                    \textbf{0.778*} &
                    0.765\hphantom{*} \\
            & & &
                2015-2016  & 
                    0.699\hphantom{*} &
                    0.775\hphantom{*} &
                    0.773\hphantom{*} &
                    0.777\hphantom{*} &
                    0.772\hphantom{*} &
                    \textbf{0.783*} &
                    0.772\hphantom{*} \\
            & & 2012-2014 &
                2014-2015  & 
                    0.778\hphantom{*} &
                    0.779\hphantom{*} &
                    0.778\hphantom{*} &
                    0.781\hphantom{*} &
                    0.783\hphantom{*} &
                    0.781\hphantom{*} &
                    \textbf{0.786}\hphantom{*} \\
            & & &
                2015-2016  & 
                    0.788\hphantom{*} &
                    0.787\hphantom{*} &
                    0.787\hphantom{*} &
                    0.789\hphantom{*} &
                    \textbf{0.792}\hphantom{*} &
                    0.789\hphantom{*} &
                    0.791\hphantom{*} \\
            & & 2012-2015 &
                2015-2016  & 
                    0.799\hphantom{*} &
                    0.800\hphantom{*} &
                    0.800\hphantom{*} &
                    0.800\hphantom{*} &
                    \textbf{0.806}\hphantom{*} &
                    0.802\hphantom{*} &
                    \textbf{0.806}\hphantom{*} \\ \midrule
    \multirow{16}{*}{\rotatebox[origin=c]{90}{\textbf{Reddit}}} &
        \multirow{3}{*}{\begin{tabular}{@{}c@{}}Topic\\Restricted\\Text\end{tabular}} & 
            2016-2017 &
                2017-2018  & 
                    0.659\hphantom{*} &
                    \textbf{0.662}\hphantom{*} &
                    0.661\hphantom{*} &
                    0.660\hphantom{*} &
                    0.660\hphantom{*} &
                    0.661\hphantom{*} &
                    0.661\hphantom{*} \\
            & & &
                2018-2019  & 
                    \textbf{0.670}\hphantom{*} &
                    0.669\hphantom{*} &
                    0.668\hphantom{*} &
                    0.668\hphantom{*} &
                    0.668\hphantom{*} &
                    0.666\hphantom{*} &
                    0.668\hphantom{*} \\
            & & &
                2019-2020  & 
                    \textbf{0.670*} &
                    0.665\hphantom{*} &
                    0.663\hphantom{*} &
                    0.665\hphantom{*} &
                    0.667\hphantom{*} &
                    0.666\hphantom{*} &
                    0.667\hphantom{*} \\
            & & 2016-2018 &
                2018-2019  & 
                    0.667\hphantom{*} &
                    0.672\hphantom{*} &
                    0.671\hphantom{*} &
                    0.671\hphantom{*} &
                    0.669\hphantom{*} &
                    \textbf{0.674}\hphantom{*} &
                    0.669\hphantom{*} \\
            & & &
                2019-2020  & 
                    0.672\hphantom{*} &
                    0.674\hphantom{*} &
                    0.674\hphantom{*} &
                    0.674\hphantom{*} &
                    0.674\hphantom{*} &
                    \textbf{0.675}\hphantom{*} &
                    0.674 \\
            & & 2016-2019 &
                2019-2020  & 
                    0.667\hphantom{*} &
                    0.668\hphantom{*} &
                    0.669\hphantom{*} &
                    0.668\hphantom{*} &
                    0.668\hphantom{*} &
                    \textbf{0.674*} &
                    0.670\hphantom{*} \\ \cmidrule{2-11}
        & \multirow{1}{*}{\begin{tabular}{@{}c@{}}SMHD\end{tabular}} & 
            2013-2014 &
                2014-2015  & 
                    0.799\hphantom{*} &
                    0.798\hphantom{*} &
                    \textbf{0.803}\hphantom{*} &
                    0.799\hphantom{*} &
                    0.799\hphantom{*} &
                    0.799\hphantom{*} &
                    0.799\hphantom{*} \\
            & & &
                2015-2016  & 
                    0.801\hphantom{*} &
                    0.800\hphantom{*} &
                    0.800\hphantom{*} &
                    \textbf{0.805}\hphantom{*} &
                    0.801\hphantom{*} &
                    0.802\hphantom{*} &
                    0.802\hphantom{*} \\
            & & &
                2016-2017  & 
                    0.792\hphantom{*} &
                    0.792\hphantom{*} &
                    0.793\hphantom{*} &
                    0.798\hphantom{*} &
                    0.797\hphantom{*} &
                    0.792\hphantom{*} &
                    \textbf{0.799}\hphantom{*} \\
            & & &
                2017-2018  & 
                    0.799\hphantom{*} &
                    0.800\hphantom{*} &
                    0.800\hphantom{*} &
                    0.803\hphantom{*} &
                    0.804\hphantom{*} &
                    0.804\hphantom{*} &
                    \textbf{0.808}\hphantom{*} \\
            & & 2013-2015 &
                2015-2016  & 
                    0.797\hphantom{*} &
                    0.795\hphantom{*} &
                    0.798\hphantom{*} &
                    0.799\hphantom{*} &
                    0.798\hphantom{*} &
                    \textbf{0.801}\hphantom{*} &
                    0.799\hphantom{*} \\
            & & &
                2016-2017  & 
                    0.786\hphantom{*} &
                    0.785\hphantom{*} &
                    0.787\hphantom{*} &
                    0.790\hphantom{*} &
                    0.790\hphantom{*} &
                    0.788\hphantom{*} &
                    \textbf{0.791}\hphantom{*} \\
            & & &
                2017-2018  & 
                    0.796\hphantom{*} &
                    0.796\hphantom{*} &
                    0.802\hphantom{*} &
                    0.799\hphantom{*} &
                    0.804\hphantom{*} &
                    0.804\hphantom{*} &
                    \textbf{0.807}\hphantom{*} \\
            & & 2013-2016 &
                2016-2017  & 
                    0.790\hphantom{*} &
                    0.790\hphantom{*} &
                    0.791\hphantom{*} &
                    0.792\hphantom{*} &
                    0.793\hphantom{*} &
                    0.792\hphantom{*} &
                    \textbf{0.794}\hphantom{*} \\ 
            & & &
                2017-2018  & 
                    0.798\hphantom{*} &
                    0.796\hphantom{*} &
                    0.804\hphantom{*} &
                    0.798\hphantom{*} &
                    0.804\hphantom{*} &
                    0.806\hphantom{*} &
                    \textbf{0.808}\hphantom{*} \\
            & & 2013-2017 &
                2017-2018  & 
                    0.799\hphantom{*} &
                    0.797\hphantom{*} &
                    0.804\hphantom{*} &
                    0.800\hphantom{*} &
                    0.803\hphantom{*} &
                    0.808\hphantom{*} &
                    \textbf{0.810}\hphantom{*} \\ \bottomrule
    \end{tabular}
    \caption{Average F1 score for the best performing vocabulary size of each feature selection method. Bolded values indicate top performers within each test set, while asterisks (*) indicate significant improvement over alternative classes of feature selection (i.e., Naive vs. Statistical vs. Semantic). Semantically-informed vocabulary selection matches or outperforms alternatives in nearly all instances, despite lacking knowledge of target outcome.}
    \label{tab:performance}
\end{table*}

\section{Predictive Generalization} \label{sec:Generalization}

Our ultimate goal is to understand how the presence of semantic shift affects downstream outcomes obtained from longitudinal analyses of social media data. Critical to the success of this goal is a methodology for controlling a statistical classifier's access to semantically unstable features when making inferences on unseen data. In this initial experiment, we demonstrate that \citeauthor{gonen2020simple}'s method for measuring semantic shift \cite{gonen2020simple} can be adapted with minimal effort to curate vocabularies with constrained levels of semantic stability. Further, we demonstrate that these vocabularies often improve predictive generalization and outperform alternative feature selection methods despite lacking an explicit awareness for the target classification outcome.

\subsection{Methods}

We design our experiment with the intention of replicating a standard deployment paradigm seen within longitudinal analyses. Language classifiers are fit on historical accumulations of annotated data and evaluated iteratively within future one-year-long time windows (see Table \ref{tab:performance}). The influence of semantic shift on generalization is measured by comparing predictive performance (F1-score) of classifiers trained using a subset of semantically-stable terms to performance of classifiers trained using alternative feature selection methods which lack awareness of semantic shift altogether. We outline the full experimental design in Appendix \ref{apx:generalization}.

\textbf{Feature Selection} Semantic stability scores $S$ are computed for each source (training) and target (evaluation) time period combination using \citeauthor{gonen2020simple}'s method \cite{gonen2020simple}. We vary vocabulary size in linear, 10-percentile intervals until all available tokens are used for training the language classifier. All vocabulary selection methods are enumerated below, chosen to encompass a variety of common strategies (na\"{i}ve and statistical) for reducing dimensionality and enhancing model performance.

\begin{itemize}
    \item \textbf{Cumulative}: Frequency $>$ 50 in the source time period
    \item \textbf{Intersection}: Frequency $>$ 50 in the source \& target time periods
    \item \textbf{Frequency}: Top $p\%$ of $n$-grams with highest frequency
    \item \textbf{Random}: Randomly selected terms, $p\%$ of the total available vocabulary
    \item \textbf{Chi-Squared}: Top $p\%$ of $n$-grams with highest chi-squared test statistic \cite{pedregosa2011scikit}
    \item \textbf{Coefficient}: Top $p\%$ of $n$-grams with highest absolute logistic regression weight within the training data
    \item \textbf{Overlap}: Top $p\%$ of $n$-grams with the highest semantic stability score $S$
    \item \textbf{Weighted (Overlap)}: Top $p\%$ of $n$-grams with the highest 50/50 weighted combination of Coefficient and Overlap scores
\end{itemize}

All feature selection methods below \emph{Frequency} (inclusive) are a subset of the \emph{Intersection} method. We introduce \emph{Weighted (Overlap)} to balance the predictive value of a given feature and its semantic stability, theorizing that a vocabulary based solely on semantic stability may come at the cost of significant predictive power, while a vocabulary solely based on within-domain predictive power will be vulnerable to generalization issues.

\subsection{Results}

\subsubsection{Stability Analysis}

To validate our implementation of \citeauthor{gonen2020simple}'s method \cite{godinic2020effects} and build context for our classification results, we first manually inspect a sample of learned semantic stability scores for each dataset. On a distribution level, we see that stability scores tend to decrease as the gap between training and evaluation time periods increases -- evidence of increased semantic shift over time. Additionally, we note that semantic stability scores within the Twitter datasets are generally lower than scores within than the Reddit datasets. These platform-specific differences align with our prior understanding of each platform's design, with Twitter tending to foster conversations motivated by current events (i.e., personal and global conflict) and Reddit offering individuals an opportunity to connect through shared interests that evolve over longer time periods \cite{noble2021semantic}. 

For all datasets, common nouns and verbs make up the majority of terms with the highest semantic stability scores (e.g., eat, bring, give, city, room, pain). These types of tokens arise only infrequently within the lower tier of semantic stability scores, typically a result of isolated conflation with current events/pop culture -- names of video games (e.g., blackout, warzone), television characters and celebrities (e.g., sandy, gore, rose), and athletic organizations (e.g., twins, braves, cal). Hashtags are frequently found in the lower semantic stability tier for the Twitter datasets, a reflection of the diversity of conversations in which they are used. Broadly, most of the observed semantic shift can be described as changes in the popularity of different word senses \cite{haase2021scot}. Although this suggests that contextual language models \cite{devlin2018bert} would be well-suited for mitigating the effect of semantic shift in longitudinal analyses, emerging research suggests this is not necessarily true in the absence of additional tuning \cite{dhingra2021time,loureiro2022timelms}.

\subsubsection{Generalization} 

We find that classifiers trained using vocabularies derived with a knowledge of semantic stability achieve equal or better predictive performance than alternative feature selection techniques in the majority of classification settings (Table \ref{tab:performance})\footnote{We exclude the Random method to save space, but note it performed significantly worse across all settings as expected.}. Semantic stability tends to be more useful for generalization within the Twitter datasets than the Reddit datasets, likely due to the aforementioned platform-specific distributions. In all cases, joint use of semantic stability and coefficient weights to derive feature selection scores (i.e., \emph{Weighted (Overlap)}) matches or moderately improves performance over use of coefficient weights in isolation. 

Finally, we note that the semantically-informed vocabulary selection methods not only offer reasonably wide operating windows (usually 20 to 50\% of the total vocabulary size), but also tend to correlate with performance within source time periods. This latter detail suggests that semantically-stable vocabulary selection can be adequately performed in the absence of validation samples from a target time period, a necessity for most longitudinal analyses. We leave hyperparameter optimization for this methodology as an area for future exploration.



\section{Practical Effects of Semantic Shift} \label{sec:ApplyExp}

\begin{figure*}[t!]
    \centering
     \includegraphics[width=0.9\linewidth]{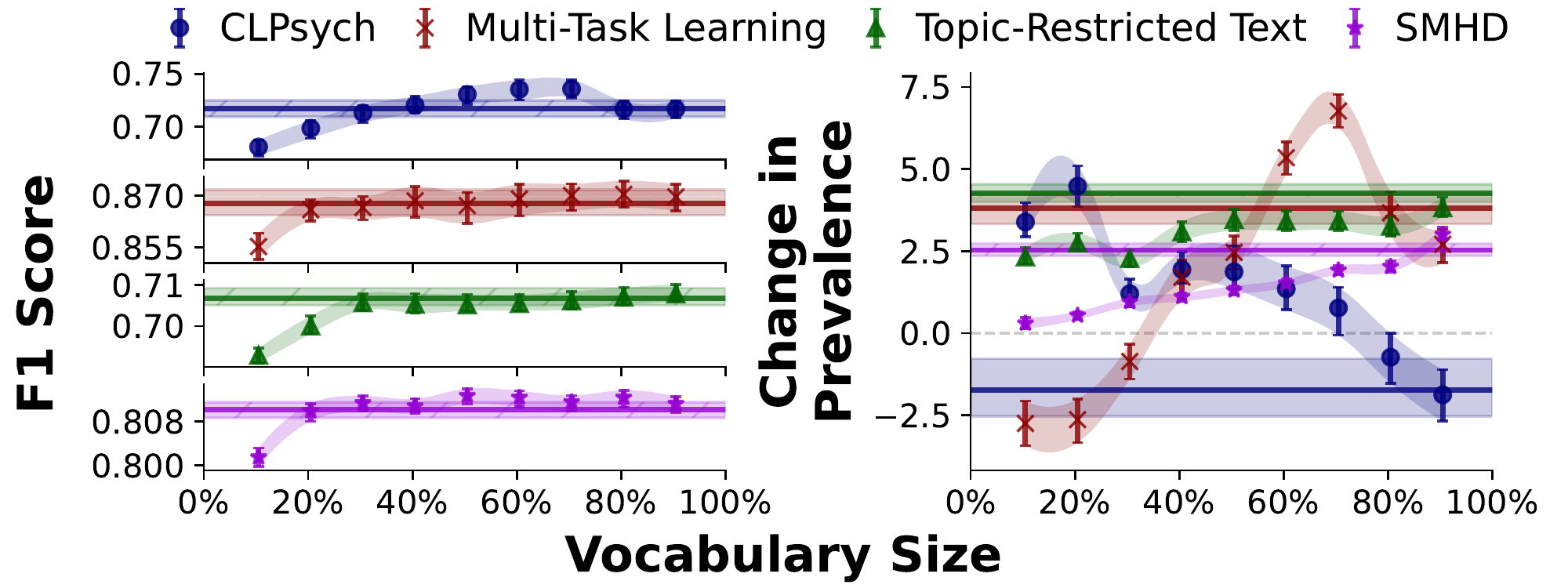}
     \caption{Horizontal bars denote each dataset's estimate under the na\"{i}ve, \emph{Intersection} baseline. Curves denote performance over varying sizes of vocabulary selected based on semantic stability $S$. (Left) Average F1 score within held-out samples drawn from each dataset's complete time period. Performance is largely indistinguishable for several of the vocabulary sizes. (Right) Estimated change in depression prevalence as a function of vocabulary.}%
     \label{fig:OutcomesSemantics}
\end{figure*}

Having demonstrated that semantically-aware vocabulary selection methods achieve comparable performance to alternative techniques using a fraction of features and can even improve predictive generalization outright, we turn our attention to understanding the practical effects semantic shift has in longitudinal modeling applications. Specifically, we leverage our ability to systematically constrain a language classifier's access to semantically volatile terms to evaluate how estimates of depression prevalence vary in the presence of semantic shift. Ultimately, we find that small changes in the vocabulary of a language classifier can promote large deviations in downstream outcomes, despite offering little to no indication of concern within historical data samples.

\subsection{Methods}

We leverage a similar experimental design to that from \S \ref{sec:Generalization}, making small methodological changes to acutely focus on understanding the practical effect of semantic shift in a deployment scenario. For example, we now model the entire time span of each annotated dataset without additional temporal splitting. Furthermore, semantic vocabulary selection is performed using embeddings learned from pairs of labeled and unlabeled datasets (e.g., CLPsych and the 1\% Twitter Stream) instead of discrete time window pairs within each of the labeled datasets. The full list of modifications is provided in Appendix \ref{apx:practical}; we release our code to support future research and allow others to reproduce our analysis.\footnote{\url{https://github.com/kharrigian/semantic-shift-websci-2022}}

To control for seasonal effects, we focus on estimating the year-over-year change in the prevalence of depression-indicative language amongst individuals in each of the unlabeled social media samples. Each unlabeled data sample is split into two distinct time periods -- March 1, 2019 to July 1, 2019 (Pre-Pandemic) and March 1, 2020 to July 1, 2020 (During-Pandemic). Classifiers are applied to each individual in the unlabeled temporal samples who meets a minimum post volume criteria -- 200 for Twitter and 100 for Reddit. We compute the prevalence of depression as the proportion of users in the unlabeled sample who have a predicted probability of depression greater than 0.5. We then measure the difference in estimated prevalence between the two time periods as a function of the underlying model vocabulary.

\subsection{Results}

\subsubsection{Language Analysis}

Semantic stability scores for each of the raw data samples (pre/during COVID-19) align with intuition regarding sociopolitical events of the era. Many of the $n$-grams with the lowest semantic stability are related to the pandemic: ``viral,'' ``masks,'' ``transmission,'' ``isolation,'' ``zoom,'' ``lockdown.'' Of terms that over-index in historical usage amongst individuals with depression, the least semantically stable include: ``panic,'' ``cuts,'' ``isolated,'' ``strain,'' ``vulnerable,'' and ``doctors.'' Each of these terms becomes closely aligned with pandemic-related phenomena that is not explicitly linked to mental-health. We provide an overview of the changes in Table \ref{tab:contextshift}; additional examples are included in Table \ref{tab:NeighborExamples} (see appendix).

\begin{table}[b]
    \centering
    \begin{tabular}{l l l}
        \textbf{Term} & \textbf{2019 Context} & \textbf{2020 Context} \\ \toprule
        Panic & Emotion (i.e., Fear) & Panic Buying, Misinformation \\ \midrule
        Cuts & Physical  & Economic \\ \midrule
        Isolated & Feeling Detached & Quarantine \\ \midrule
        Strain & Discomfort/Pressure & Virus \\ \midrule
        Vulnerable & Emotion & At-risk Populations \\ \midrule
        Doctors & Personal Experience & Frontline Workers, PPE \\ \bottomrule
    \end{tabular}
    \caption{Change in the most prevalent context from 2019 to 2020 for a handful of terms which historically over-indexed in usage amongst individuals living with depression.}
    \label{tab:contextshift}
\end{table}



\subsubsection{Prevalence Estimates}

Turning our attention to the statistical classifiers, we observe that predictive performance as a function of the underlying vocabulary is nearly indistinguishable for vocabularies of size 40\% and higher. However, as shown in Figure \ref{fig:OutcomesSemantics}, we identify significant differences in the estimated change in population-level depression prevalence as a function of the model's underlying vocabulary. In some cases, these differences are relatively minor and lead to the same general conclusions. In other cases, we arrive at entirely different statements regarding the directional change of depression prevalence (i.e., increase instead of decrease) and absolute change (i.e., nearly 10\% in the case of the minimum and maximum Multi-Task Learning estimates). 



\section{Ethical Considerations} \label{sec:Ethics}

Population monitoring at scale warrants an ethical discussion. We discuss some trade-offs between risk and reward, specifically when studying sensitive characteristics (e.g., mental health status) from social media. We direct the reader to work from \citet{conway2016social} and \citet{golder2017attitudes} for an expanded review.

\subsubsection{Risks} Two serious risks arise from measuring personal characteristics using social media data: 1) discrimination, and 2) measurement error. The former is a challenge associated with any approach used to acquire information about human characteristics or behavior, whether inferred by an algorithm or not. Knowledge of personal attributes could be used by educational institutions to make biased admissions decisions, by law enforcement to track individuals without cause, or by political/government entities to target vulnerable individuals. These concerns are particularly poignant with regards to stigmatized characteristics, such as mental illness. Discriminatory actions based on these characteristics could have long-lasting financial and social consequences -- e.g., difficulty obtaining loans, increased insurance premiums, and exclusion from certain communities. While statistical models are not the only method for gathering this information, they can be used in some situations where other approaches are infeasible \cite{paul2016social}. With respect to the second challenge, we draw the reader's attention to substantial evidence that demonstrates language models trained on social media datasets perform disproportionately amongst different demographic groups \cite{aguirre2021genderandracial} and maintain historical social biases \cite{brunet2019understanding}. These systematic errors in models of mental health may further exacerbate social stratification in a opaque and elusive manner \cite{bender2021dangers}.

\subsubsection{Rewards} We must also take care not to ignore the tremendous need for these methods and the benefits they bring. The same technology used to ostracize vulnerable individuals could also be used to provide those individuals with social services. Likewise, access to reasonably accurate classifiers with well-defined bounds of uncertainty could help small organizations acquire sufficient data to optimize resource allocation without needing to invest in the cost-prohibitive infrastructure necessary to execute traditional monitoring at scale (e.g., random digit dialing, online surveys) \cite{vaske2011advantages,shaver2019using}. These opportunities come with a variety of additional advantages over traditional population monitoring mechanisms -- social media monitoring preempts the need to use downstream outcomes that are not useful in situations that require immediate decisions (e.g., latent changes in suicide rate), addresses certain forms of sample bias (e.g., selection bias introduced when individuals opt into a survey, disclosure bias that emerges when individuals are hesitant to discuss stigmatized topics with an interviewer), and provides the opportunity to make comparisons against retrospective baselines. Moreover, a significant amount of work focuses on methods to mitigate risk of discrimination \cite{zhao2018gender} and adequately correct for sampling biases specific to social media data \cite{giorgi2021correcting}. The large body of literature on social media monitoring in public health, for example, evidences the tremendous need for these technologies \cite{paul2017social}. It is our responsibility to develop and deploy them in an ethically responsible manner.

\subsubsection{Discussion} Practitioners must weigh these trade-offs in the context of their particular use case. In our use case, we note the goal of this study is \emph{not} to make claims about a particular longitudinal trend or even demonstrate the prowess of a statistical modeling approach. Rather, our intention is to understand whether existing models can be trusted for measuring longitudinal trends at all in the presence of semantic shift, and if not, identify potential opportunities for practitioners to improve reliability of their models. The utility of such an exploration would be questionable if these types of models had not already been deployed in academia and beyond. However, one need only to look at research published within the last year regarding COVID-19 to see that machine learning classifiers are actively being used to understand a variety of social dynamics, ranging from mental health outcomes \cite{fine2020assessing,tabak2020temporal} to transportation usage \cite{morshed2021impact}. These analyses will form a foundation for public policy in the coming post-pandemic years. It is critical that we answer: are these results reliable?


\section{Discussion} \label{sec:Discussion}

In this study, we demonstrated that semantic shift can be problematic in longitudinal monitoring applications, both in terms of pure predictive performance and our ability to estimate population-level outcomes. The method for measuring semantic stability introduced by \citet{gonen2020simple} and adapted for use as a feature selection method here is promising for reducing domain divergence and improving generalization over time. However, more research must be done to understand which deployment scenarios may obtain the most significant benefit from its use.

\subsubsection{Limitations} The outcomes of this study are designed to spur conversation amongst practitioners, not necessarily to provide a panacea for addressing semantic noise in deployment scenarios. Indeed, we recognize our quantitative experiments are limited by the annotated data itself. For example, it remains to be seen whether semantically stable vocabularies are most useful over the course of certain time frames (e.g., decades instead of years), within a subset of social media platforms, or in the context of specific modeling tasks. Moreover, the labeled datasets may not be entirely conducive to the longitudinal classification experiments we performed in \S \ref{sec:Generalization} \cite{demasi2017meaningless,tsakalidis2018can}, with depression known to present episodically within individuals \cite{collishaw2004time,angst2009long}.  Given these dataset constraints, we urge researchers to consider replicating our analysis using new datasets which feature different underlying temporal dynamics.

\subsubsection{Next Steps} Beyond expanding the analysis to a more diverse array of datasets and target measures, we foresee substantial value in continued exploration of semantic stability's effect on predictive generalization under alternative technical perspectives. For example, we note that our current study focuses solely on discrete time windows, an abstraction that is useful for simple monitoring applications, but too constraining for others. It would be of significant value to the longitudinal monitoring community to evaluate whether continuous time and diachronic embeddings offer advantages over their discretized counterparts \cite{hamilton2016diachronic,huang2019neural}. We also recognize that our implementation operates in two distinct stages (i.e., feature selection, model training), a setup which may inhibit performance. A better approach may involve leveraging knowledge of semantic shift to explicitly regularize coefficients at training time.


\bibliographystyle{ACM-Reference-Format}
\bibliography{main}


\appendix

\section{Preprocessing}

To promote consistent analysis, we use automatic language identification \cite{lui2012langid} to isolate English text in each dataset (labeled and unlabeled). Additionally, per the recommendations of \citet{de2014mental}, we exclude posts in the labeled datasets that either contain a match to a mental health related $n$-gram or are drawn from a subreddit explicitly dedicated to providing mental health support. This filtering is designed to encourage statistical models to learn robust linguistic relationships with depression as opposed to those introduced by sampling-related artifacts.

\section{Training Parameters} \label{apx:predparam}

Vocabularies of a maximum 500k $n$-grams (unigrams, bigrams, and trigrams only) are learned using Gensim's Phraser module, parameterized using a PMI threshold of 10 and minimum frequency of 5 \cite{mikolov2013distributed}. For classification experiments, all posts sampled for a user from a given time period are tokenized using a Python implementation of Twokenizer \cite{o2010tweetmotif}, rephrased using the time period's specific Phraser model, concatenated together, and translated into a single document-term vector. Representations are transformed using TF-IDF weights learned at training time before being $\ell_{2}$-normalized. As a classification architecture, we use $\ell_{2}$-regularized logistic regression, optimizing parameters using limited-memory BFGS as implemented in Scikit-learn \cite{pedregosa2011scikit}. Inverse regularization strength $C$ is selected independently for each training sample such that F1 score is maximized within the development splits of a 10-fold cross validation run.

To measure semantic shift, the same vocabularies and Phrase detection models introduced above serve as the foundation for training unique Word2Vec models \cite{mikolov2013distributed} for each dataset and time period. We leverage Gensim's implementation of the continuous bag of words (CBOW) formulation of Word2Vec to learn 100-dimensional embeddings, training each model for 20 iterations using the default window and negative sampling sizes \cite{mikolov2013efficient}. We obtain semantic neighborhoods for each $n$-gram $w$ using $k = 500$, $\text{cf}_{\text{nb}}=50$, and $\text{cf}_{\text{shift}}=50$. Alternative neighborhood sizes ($k=250, 1000$) and frequency thresholds ($\text{cf}_{\text{nb}}=10, 25, 100$; $\text{cf}_{\text{shift}}=25, 100$) did not have a significant effect on downstream outcomes. In line with \citet{gonen2020simple}, we measure vector similarity using cosine distance.

\section{Generalization: Expanded} \label{apx:generalization}

\subsubsection{Experimental Design} Each instance of an experimental run consists of multiple stages. In the first stage, users are randomly allocated into a 80/20 train/test split, with data from those in the training group used to learn word embeddings for each time period within a dataset (i.e., historical accumulations and future one-year windows). During the second stage, users in both the training and test splits are independently resampled to form subsets for training and evaluating the language classifiers. This secondary sampling step is constrained such that all users meet a minimum post threshold within each time period\footnote{200 for Twitter, 100 for Reddit; Reddit comments contain 2x as many tokens on average.}, the number of users within each training time period is equivalent, and that classes are balanced in both training and test subsets. The first and second stages are repeated 100 and 10 times, respectively, providing us with 1,000 experimental samples for each dataset. Classifiers are evaluated using data from users in the 20\% test split sampled during the first stage of the experiment.

\section{Practical Effects: Expanded} \label{apx:practical}

\subsubsection{Experimental Design} In the first stage of the experimental procedure, we fit 10 embedding models for each of the labeled datasets -- using randomly sampled subsets (80\% size) of the complete dataset -- and three embedding models for each of the unlabeled data samples -- one using data from the entire Jan. 1, 2019 to July 1, 2020 time period, one using data from March 1, 2019 to July 1, 2019, and one using data from March 1, 2020 to July 1, 2020. The latter two unlabeled data models are used to qualitatively identify language which has undergone semantic shift since the start of the COVID-19 pandemic, while the former model is used in conjunction with the labeled dataset models to identify semantically stable vocabularies for training classifiers. To reduce computational expense, we randomly sample 20\% of posts to train each of the unlabeled data embedding models, but otherwise maintain the same hyperparameters and training settings enumerated in \S \ref{sec:Generalization}. The second stage of the experiment proceeds as before, resampling amongst the users allocated to the training split of the annotated data to derive semantically stable vocabularies and train language classifiers. We perform the second stage 10 times, providing us with 100 classifiers for each labeled dataset.

\subsubsection{Performance/Prevalence} 
\begin{figure}[h!]
    \centering
    \includegraphics[width=.9\columnwidth]{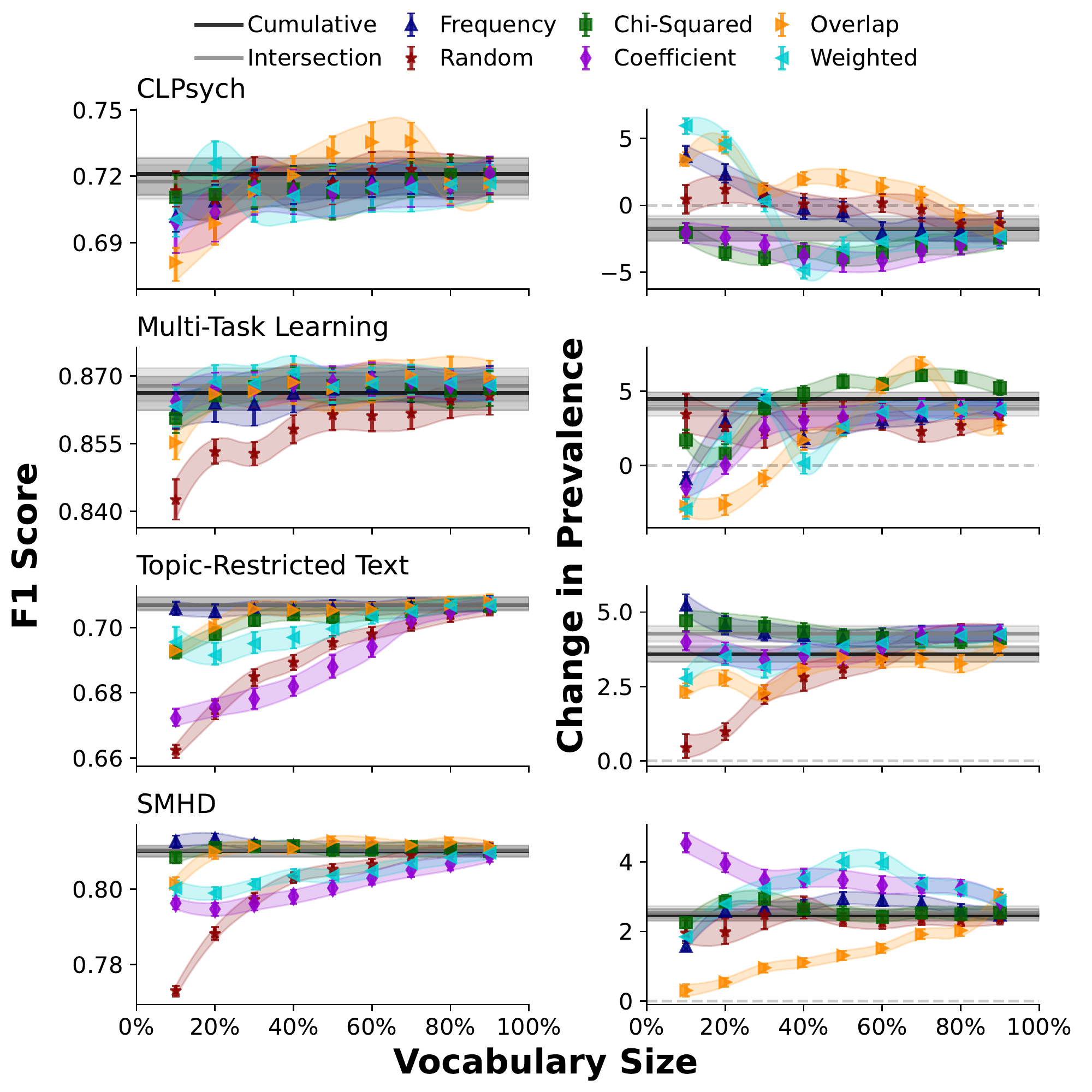}
    \caption{Comparison of within-domain predictive performance and estimated change in depression prevalence as a function of the underlying vocabulary.}
    \label{fig:prevperv}
\end{figure}

Figure \ref{fig:prevperv} shows predictive performance of classifiers within their source data domains compared to estimated change in depression prevalence for all datasets and vocabulary selection methods. We note that estimates of change in depression prevalence vary more dramatically within the Twitter data than the Reddit data. In all cases outside of training on the CLPsych Shared Task dataset \cite{coppersmith2015clpsych}, semantically-stable vocabularies tend to reduce the estimated change in prevalence. Randomly sampled vocabularies tend to achieve the lowest within-domain performance; as expected, they also generate estimated outcomes that vary less from the Cumulative and Intersection baselines compared to alternative feature selection methods.

\subsubsection{Language Analysis} We include a sample of terms from the unlabeled Twitter and Reddit data samples which underwent significant semantic shift in Table \ref{tab:NeighborExamples} (next page). It becomes clear through this analysis that the COVID-19 pandemic is a primary driver of the shift. Several terms which were previously used to describe emotional states (e.g., \emph{panic}, \emph{isolated}, \emph{relief}) are now used to describe pandemic-specific phenomena. 

\begin{table*}
    \centering
    \begin{tabular}{l l l l}
         & \textbf{Token} & \textbf{2019 Context} & \textbf{2020 Context} \\ \toprule
         \multirow{18}{*}{\rotatebox[origin=c]{90}{\textbf{Twitter}}} 
         & Panic &
            \begin{tabular}{@{}l@{}}rage, meltdown, anxiety, anger,\\barrage, migraine, phobia, outrage,\\manic, rush, asthma\end{tabular} &
            \begin{tabular}{@{}l@{}}hysteria, chaos, fear, misinformation,\\confusion, frenzy, paranoia, mayhem,\\insanity, fearmongering\end{tabular} \\  \cmidrule{2-4}
          & Eviction &
            \begin{tabular}{@{}l@{}}deportation, emergency, cancellation,\\negligence, reservation, injunction, immediate,\\ incarceration, inaction, cancellations, termination\end{tabular} &
            \begin{tabular}{@{}l@{}}evictions, repayment, foreclosure, indefinite,\\immediate, injunction, moratorium,\\visitation, bankruptcy, deportation\end{tabular}\\  \cmidrule{2-4}
         & Crisis &
            \begin{tabular}{@{}l@{}}humanitarian crisis, crises, catastrophe,\\epidemic, disaster, threat, emergency,\\issue, income inequality\end{tabular} &
            \begin{tabular}{@{}l@{}}pandemic, crises, crisi, catastrophe, \#pandemic,\\cris, pand, disaster, epidemic, outbreak,\\ pandem, pandemi, downturn\end{tabular} \\  \cmidrule{2-4}
         & Corona &
            \begin{tabular}{@{}l@{}} bourbon, margarita, mesa, distillery,\\carmel, lager, grove, hut, riverside,\\fireball, mayo, scottsdale, tempe\end{tabular} &
            \begin{tabular}{@{}l@{}}carona, covid, coro, virus, \#corona,\\rona, \#covid\_19, \#coronavirus, ebola,\\\#covid, vir, \#covid2019, \#wuhanvirus\end{tabular} \\  \cmidrule{2-4}
         & Doctors &
            \begin{tabular}{@{}l@{}}psychiatrists, medically, clinic, cps,\\accountants, police, miscarriages, malpractice,\\abortions, prescribe, counseling,
	        procedure\end{tabular} &
            \begin{tabular}{@{}l@{}}midwives, \#nurses, \#doctors, epidemiologists, emts,\\ front-line, \#coronawarriors, frontliners, virologists,\\masks, ppe, respirators, docs, heroes\end{tabular} \\  \cmidrule{2-4}
            
         & Zoom &
            \begin{tabular}{@{}l@{}}zooming, zooms, tap, log, zoomed, hop,\\hover, dial, jump, slide, camera, plugged,\\optical, infrared, roll, nikon\end{tabular} &
            \begin{tabular}{@{}l@{}}skype, webex, \#zoom, hangouts,\\\#microsoftteams, webcam, facetime,\\telephone, livestream, classroom \end{tabular} \\  \midrule
         \multirow{21}{*}{\rotatebox[origin=c]{90}{\textbf{Reddit}}} 
         & Panic &
            \begin{tabular}{@{}l@{}}rage, anger, despair, desperation, reflex,\\terror, adrenaline, silence, anxiety,\\dread, paranoia, laughter\end{tabular} &
            \begin{tabular}{@{}l@{}}hysteria, fear, panicking, paranoia,\\civil unrest, toilet paper, adrenaline, diarrhea,\\virus, corona, anxiety, shutdowns, \#panicbuying \end{tabular} \\  \cmidrule{2-4}
         & Cuts &
            \begin{tabular}{@{}l@{}}cut, jumps, runs, cutting, pulls, moves,\\bounces, falls, turns, burns, drags, dips,\\breaks, bursts, rips, goes, bumps\end{tabular} &
            \begin{tabular}{@{}l@{}}cut, cutting, subsidies, budgets,\\deductions, revenues, checks, payments,\\breaks, deals, figures, loans, deposits, gains\end{tabular} \\  \cmidrule{2-4}
         & Isolated &
            \begin{tabular}{@{}l@{}}unpleasant, unstable, detached, unsafe,\\populated, invasive, unknown, confined,\\endangered, absent, vulnerable, insulated\end{tabular} &
            \begin{tabular}{@{}l@{}}quarantined, isolating, separated, enclosed,\\insulated, infectious, confined, active, populated,\\autonomous, vulnerable, detached\end{tabular} \\  \cmidrule{2-4}
         & Vulnerable &
            \begin{tabular}{@{}l@{}}susceptible, dangerous, prone, unstable, aggressive,\\hostile, disruptive, detrimental, receptive,\\fragile, damaging, sensitive\end{tabular} &
            \begin{tabular}{@{}l@{}}susceptible, dangerous, immunocompromised, infectious,\\isolating, elderly, disadvantaged, contagious,\\tolerant, likely, isolated, symptomatic\end{tabular} \\  \cmidrule{2-4}
         & Relief &
            \begin{tabular}{@{}l@{}}satisfaction, sadness, fatigue, disappointment,\\warmth, despair, desperation, payoff,\\ excitement, dread, nausea, accomplishment\end{tabular} &
            \begin{tabular}{@{}l@{}}assistance, stimulus, aid, bailout, funding,\\compensation, temporary, medicaid, fund, bailouts,\\cheque, unemployment, benefits, donations, payout\end{tabular} \\  \cmidrule{2-4}
        & Strain & 
            \begin{tabular}{@{}l@{}} inflammation, deficiency, dose, stress, pressure,\\calcium, medication, concentration, tissue, nausea,\\receptors, doses, acne, effect, discomfort\end{tabular} &
            \begin{tabular}{@{}l@{}}disease, illness, infections, symptom, mutation,\\virus, outbreak, pneumonia, infection, strains, influenza,\\epidemic, dependency, 	    diseases, allergy, flu, coronaviruses\end{tabular} \\ \cmidrule{2-4}
         & Testing &
            \begin{tabular}{@{}l@{}}test, tests, qa, certification, training,\\scans, screening, monitoring, research, filtering,\\ tested, imaging, treatments, coding, experiments\end{tabular} &
            \begin{tabular}{@{}l@{}}tests, contact tracing, screening, test,\\containment, infections, transmission, tracing,\\ infection, ppe, ventilators, tested, lockdowns\end{tabular} \\ \bottomrule
    \end{tabular}
    \caption{Examples of terms which experienced significant semantic shift from 2019 to 2020.}
    \label{tab:NeighborExamples}
\end{table*}

\end{document}